\documentclass{article}

\usepackage{microtype}
\usepackage{graphicx}
\usepackage{subfigure}
\usepackage{booktabs} 

\usepackage[table]{xcolor} 

\usepackage{hyperref}

\usepackage{lipsum}

\usepackage[dvipsnames]{xcolor}
\def\cmark{\textcolor{ForestGreen}{\checkmark}}
\def\xmark{\textcolor{red}{\texttimes}}

\usepackage[accepted]{icml2025}

\usepackage{amsmath}
\usepackage{amssymb}
\usepackage{mathtools}
\usepackage{amsthm}

\usepackage[capitalize,noabbrev]{cleveref}

\theoremstyle{plain}

\theoremstyle{definition}

\theoremstyle{remark}

\usepackage[textsize=tiny]{todonotes}

\definecolor{mplred}{HTML}{D62728}

\usepackage[most]{tcolorbox} 
\tcbset{boxsep=1ex,left=1em,right=1em,top=0.6ex,bottom=0.6ex,
        colback=black!3,colframe=black!35,sharp corners}
\newtcolorbox{snippetbox}[1]{title={#1 \footnotesize(snippet)}, breakable}

\icmltitlerunning{Enhancing Agentic Autonomous Scientific Discovery with Vision-Language Model Capabilities}

\begin{document}

\twocolumn[
\icmltitle{Enhancing Agentic Autonomous Scientific Discovery with Vision-Language Model Capabilities}



\icmlsetsymbol{equal}{*}

\begin{icmlauthorlist}
\icmlauthor{Kahaan Gandhi}{Caltech,Cambridge,Haverford}
\icmlauthor{Boris Bolliet}{Cambridge,Kavli}
\icmlauthor{Íñigo Zubeldia}{Kavli,IoA}

\end{icmlauthorlist}

\icmlaffiliation{Cambridge}{Department of Physics, University of Cambridge, Cambridge, United Kingdom}
\icmlaffiliation{Kavli}{Kavli
Institute for Cosmology, University of Cambridge, Cambridge,
United Kingdom}
\icmlaffiliation{Haverford}{Department of Physics and Astronomy, Haverford College, 370 Lancaster Avenue, Haverford, PA 19041, USA}
\icmlaffiliation{Caltech}{Division of Physics, Mathematics
and Astronomy, California Institute of Technology, Pasadena, CA
91125, USA}
\icmlaffiliation{IoA}{Institute of Astronomy,
University of Cambridge, Cambridge, United Kingdom}

\icmlcorrespondingauthor{Kahaan Gandhi}{kahaan@gmail.com}

\icmlkeywords{Machine Learning, ICML}

\vskip 0.3in
]



\printAffiliationsAndNotice{}  

\begin{abstract}
We show that multi-agent systems guided by vision-language models (VLMs) improve end-to-end autonomous scientific discovery. By treating plots as verifiable checkpoints, a VLM-as-a-judge evaluates figures against dynamically generated domain-specific rubrics, enabling agents to correct their own errors and steer exploratory data analysis in real-time. Case studies in cosmology and astrochemistry demonstrate recovery from faulty reasoning paths and adaptation to new datasets without human intervention. On a 10-task benchmark for data-driven discovery, VLM-augmented systems achieve pass@1 scores of 0.7–0.8, compared to 0.2–0.3 for code-only and 0.4–0.5 for code-and-text baselines, while also providing auditable reasoning traces that improve interpretability.

\end{abstract}

\section{Introduction}

Autonomous discovery systems are beginning to solve research tasks once thought to require human expertise. Large language model (LLM) agents can already propose hypotheses \cite{si2024llmsgeneratenovelresearch, su-etal-2025-many}, retrieve relevant literature \cite{Baek2025ResearchAgent, agarwal2024litllm}, execute computational experiments \cite{Boiko2023, CMBAGENT_2025, xu2025opensourceplanning}, and even draft papers \cite{villaescusanavarro2025denarioprojectdeepknowledge,Denario_2025,CMBAGENT_2025}. In principle, LLMs can be applied across the research pipeline, but not all tasks are equally verifiable. For example, code generation and mathematical derivations are objective and therefore easier to evaluate, while visual reasoning and communication are more subjective and require discretion. When orchestrated into multi-agent systems for end-to-end automation, these harder-to-verify tasks often emerge as failure modes. 

For autonomous systems to become credible scientific collaborators, they must move beyond analysis alone and communicate findings in ways interpretable to the research community. In data-intensive fields, figures are the primary medium for both communication and interpretation. They compress large datasets into digestible representations while also guiding the research process: plots reveal anomalies, prompt the reconsideration of hypotheses, and steer subsequent steps. This feedback loop is central to human discovery workflows but remains largely absent in current systems, where frontier models fail to handle domain-specific conventions in plots \cite{joseph2025astrovisbench}.

To address this gap, we extend \href{https://github.com/CMBAgents/cmbagent}{\texttt{cmbagent}}, a fully autonomous discovery system \cite{Laverick:2024fyh,CMBAGENT_2025,xu2025opensourceplanning}, by introducing two new multi-agent workflows that are steered by visualization. The first is a multimodal self-correction loop, where a vision–language model (VLM) judges agent-generated plots for consistency with domain-specific physical expectations and visual clarity. The second is a scientific anomaly-detection system, where unexpected results automatically trigger exploratory experiments, comparative analyses, and realignment of the research trajectory. 

\section{Related Work}

\begin{figure*}[t]
  \centering
  \includegraphics[width=\textwidth]{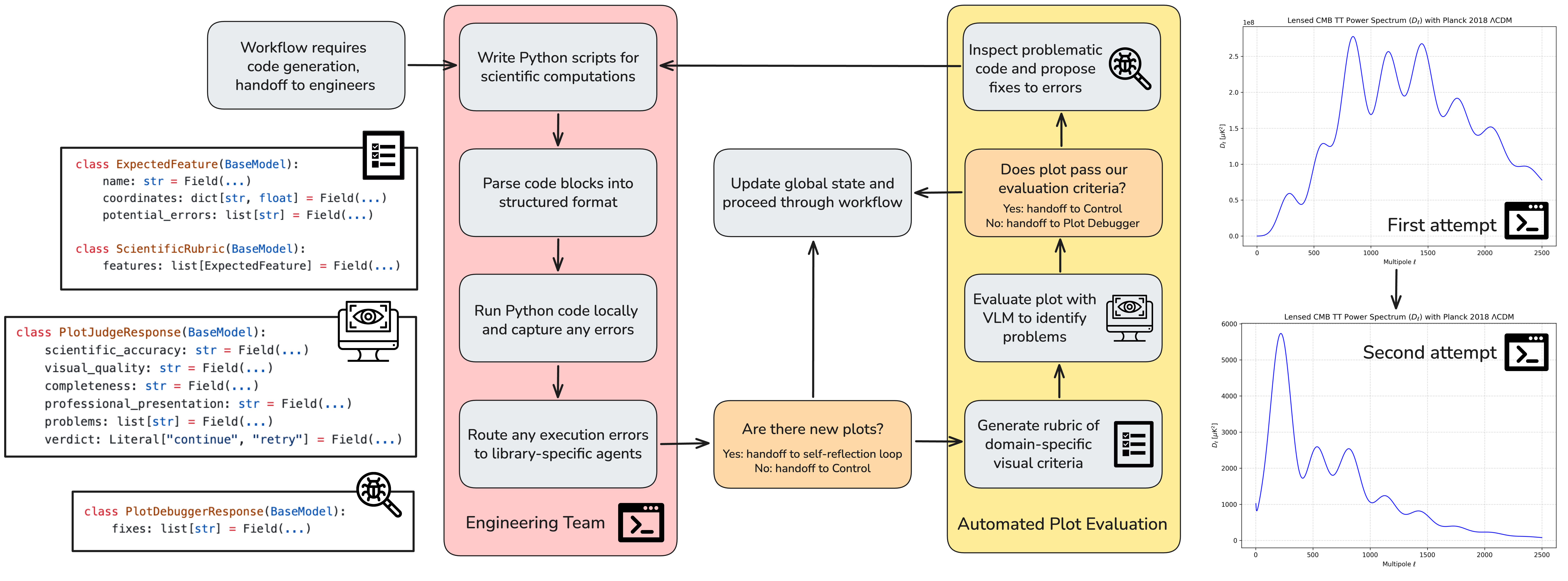}
  \caption{\textit{Self-correction workflow with a VLM-as-a-judge.} Center: feedback loop where plots are evaluated by a VLM against a domain-specific rubric, routed to the Plot Judge and Plot Debugger agents when criteria are not met, and iteratively revised. Left: Pydantic schemas defining structured outputs, with icons showing where each schema is applied. Right: first attempt by coding agents to plot the lensed CMB TT power spectrum from Sec.~\ref {sec:CMB}, where $D_\ell^{TT}$ was incorrectly rescaled, and the corrected version after a single pass, now in agreement with Planck 2018 $\Lambda$CDM predictions.}
  \label{fig:correction}
\end{figure*}

\textbf{LLM-assisted scientific visualization}. Work on LLM-assisted visualization has evolved from simple query-to-plot translation to agentic systems that iteratively generate and refine scientific figures. \texttt{LLM4Vis} \cite{wang-etal-2023-llm4vis} frames visualization as a recommendation task, prompting models with few-shot examples to suggest chart types, while \texttt{HAIChart} \cite{10.14778/3681954.3681992} applies a Monte Carlo Graph Search–based generation algorithm with reinforcement learning and human feedback to efficiently explore the visualization space. \texttt{MatPlotAgent} \cite{yang2024matplotagent} introduces a model-agnostic framework that incorporates VLMs for self-reflection, closing the feedback loop for autonomous plotting. \texttt{PlotGen} \cite{10.1145/3701716.3716888} expands this into a multi-agent system that integrates numerical, lexical, and visual feedback to iteratively refine plots across modalities. Across these approaches, domain-specific rigor is either supplied by humans in the loop or embedded in the task specification itself. In contrast, scientific workflows demand rigor that develops over longer-horizon trajectories, where plots serve as checkpoints to be interpreted in context and used to guide reasoning, rather than validated solely for visual clarity.

\textbf{Evaluating scientific visualizations}. Efforts to evaluate LLM-generated visualizations have relied either on human judgment or on automatic methods that approximate it. \texttt{MatPlotBench} \cite{yang2024matplotagent} modifies tasks from Matplotlib and OriginLab galleries, using a VLM-as-a-judge with high correlation to human-annotated scores. \texttt{VisEval} \cite{10.1109/TVCG.2024.3456320} recognizes that different LLMs prefer different ways of describing data, and introduces generalized metrics focused on readability, validity, and legality rather than strict one-to-one correctness. Both benchmarks emphasize the generation of precise plots using scientific libraries, typically framed as shorter-horizon, well-defined tasks with fixed gallery targets. In contrast, scientific discovery is more ambiguous: tasks evolve iteratively, and correctness depends on reasoning in a domain-specific context. \texttt{AstroVizBench} \cite{joseph2025astrovisbench} targets astronomy-specific workflows by distilling expert-written notebooks into processing and visualization tasks, again maximizing correlation between VLMs and human evaluators. But, the setup requires multiple layers of ground truth, including setup and processing code, the final visualization, and underspecifications that supplement the natural language query (e.g., providing the correct filters, axis limits, or marker sizes). Such requirements do not apply to online evaluation settings, where correction must occur in real time without access to any ground truth. \texttt{ScienceAgentBench} \cite{chen2024scienceagentbenchrigorousassessmentlanguage} requires agents to generate self-contained Python programs that produce specified artifacts, with visualization quality included as one of several metrics (alongside execution, task success, semantic similarity, and cost). In this setting, however, figures serve only as the final deliverable, with no role in shaping the intermediate analysis.

\textbf{Autonomous discovery systems}. Several recent systems aim to automate broader segments of the scientific process. The \texttt{AI Scientist-v1} \cite{lu2024aiscientist} partitions research into three phases (idea generation, experiment iteration, and paper-writing), recording the outcome of each experiment but conditioning exploration only on text. The \texttt{AI Scientist-v2} \cite{aiscientist_v2} retains this structure while adding agentic tree-search-based exploration during the experiment phase, where VLMs improve clarity and flag bugs as faulty nodes. Here, vision informs search by correcting existing errors rather than pursuing new directions. Notably, this system produced the first fully AI-generated paper to pass peer review. \texttt{CodeScientist} \cite{jansen2025codescientistendtoendsemiautomatedscientific} addresses the challenge of distinguishing errors from potential discoveries by adopting a semi-automated workflow, keeping a human in the loop to select promising ideas and verify results. \texttt{Self-Debugging} \cite{chen2024teaching} incorporated self-reflection, showing that LLMs can iteratively debug code by explaining and revising outputs, but it remained limited to programming tasks where success can be judged solely by the correctness of the final state. More recently, \texttt{Denario} \citep{villaescusanavarro2025denarioprojectdeepknowledge,Denario_2025,CMBAGENT_2025} applied these ideas to actual scientific datasets spanning many disciplines, with one full-AI generated paper accepted at the \href{https://agents4science.stanford.edu/}{Open Conference of AI Agents for Science 2025}. It is worth noting that the level at which these systems operate is already manifestly on par with that of domain expert human researchers\footnote{In a  NeurIPS Challenge 2025, a team using \texttt{cmbagent} topped \href{https://www.codabench.org/competitions/8934/\#/results-tab}{the leaderboard} of the open-submission phase of the competition.}.

\section{Methodology}

\subsection{System Overview}

\texttt{cmbagent} follows a planning and control strategy: \textbf{Planner} and \textbf{Plan Reviewer} agents iteratively propose and critique a plan that breaks the user’s task into a sequence of subtasks. The finalized plan is passed to the \textbf{Control} agent, which executes the plan step-by-step by sequentially handing subtasks to specialized agents. The orchestration, handoffs, and message routing are implemented with AG2. Specialized agents only receive pruned subsets of the shared conversational state, while the Control agent maintains access to the complete history. Code generation is handled by an engineering team of four agents, which uses AG2’s nested chats to generate code, structure it into executable blocks, run those blocks locally, and route errors. The \textbf{Researcher} agent can provide discussion or background, supplementing the generated code with contextual information. Some handoffs, including those within the engineering team, are rigidly enforced to preserve deterministic workflows, whereas in other cases, Control dynamically selects the most relevant agent.

We apply checkpointed evaluations around the code generation step. Intermediate steps in scientific workflows may be valid or invalid depending on the broader trajectory, making it difficult to judge progress step-by-step. However, evaluating only end states also obscures where errors arise within a chain of agent decisions. Instead, we treat the appearance of a new plot as a checkpoint: bounded by known scientific constraints and therefore well-suited for feedback loops and intermediate self-reflection. When the engineering team generates a new plot, a GPT-4o model creates a domain-specific rubric of visual criteria, decomposing the scientific requirements of the natural language prompt into a set of checkable items. This rubric is designed to resemble the (text, image) pairs that VLMs are trained on to provide a layer of automatic prompt engineering. The resulting plot is then subjected to a black-box test, judged solely against the rubric without reference to the underlying code, to assess whether the series of subtasks leading to the figure was successfully executed.

The system is configured to run in one of two distinct modes, which determine the handoffs following the plot checkpoint. In the correction setting, if the VLM judges that a plot does not satisfy the scientific rubric, execution is routed into a workflow where the figure is debugged and revised until the criteria are met. In the discovery setting, if the VLM instead highlights features that appear scientifically relevant, \texttt{cmbagent} invokes a workflow for exploratory data analysis, using those features as guiding signals and adapting the overall research trajectory based on the intermediate findings. 

\subsection{Automating Self-Correction with VLM Feedback}

\begin{figure*}[t]
  \centering
  \includegraphics[width=\textwidth]{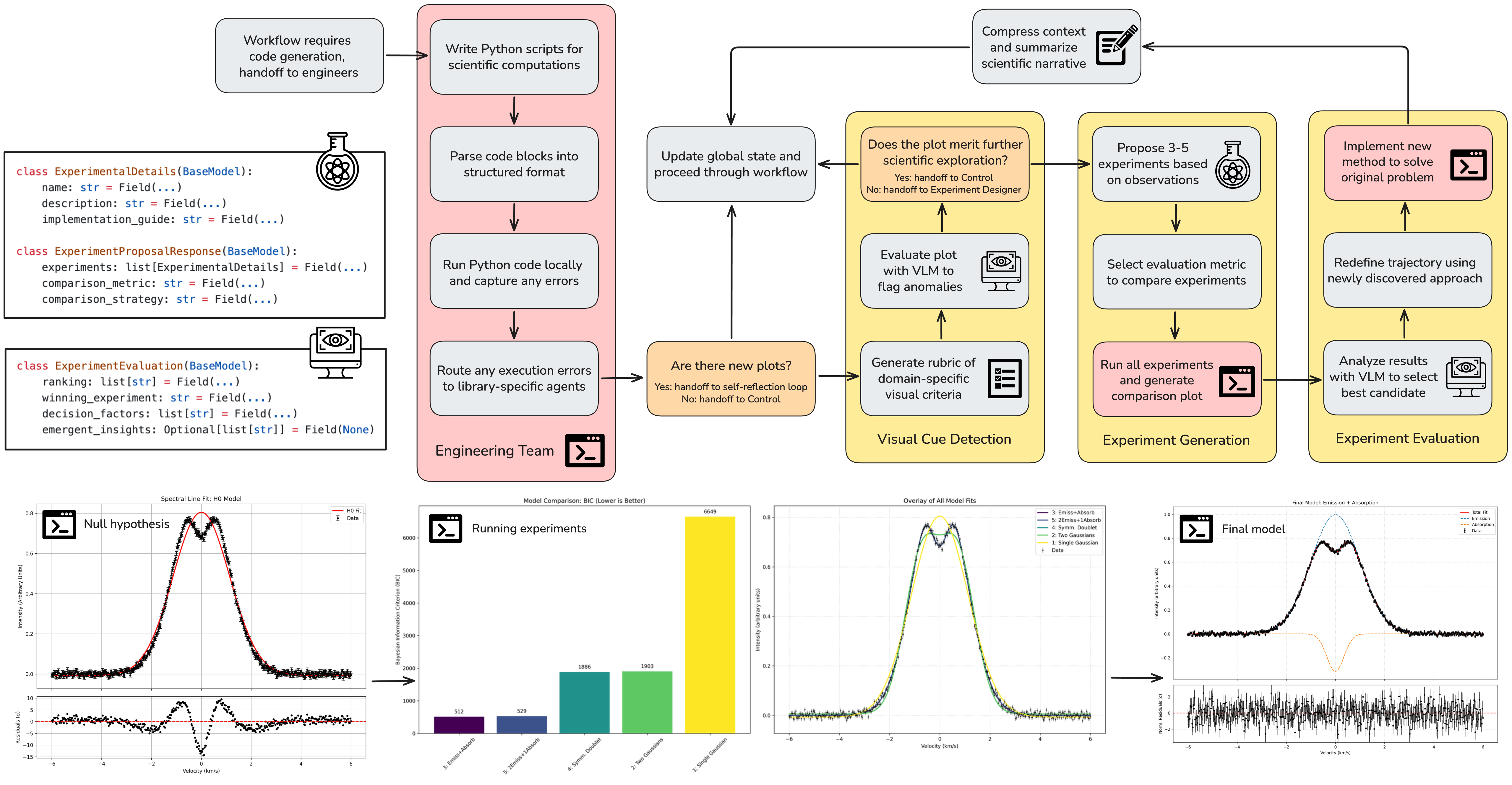}
    \caption{\textit{Extending scientific workflows with VLM-guided exploration.} Top right: when plots reveal features warranting further investigation, the system initiates exploratory data analysis to adapt the research trajectory and update beliefs with intermediate findings. Top left: Pydantic schemas define structured outputs for experiment design and evaluation, in addition to those shown in Fig.~\ref{fig:correction}. Bottom: three phases of the case study in Sec.~\ref{sec:CaseStudy} — (1) fitting the null hypothesis of a single Gaussian, (2) testing alternative spectral line models, and (3) selecting the self-absorption model, correctly inferring the true distribution of the data.}  \label{fig:discovery}
\end{figure*}

When the goal is to catch and correct faulty plots in real time, the engineering team is instructed to supplement the original task with visual checks (e.g., residual subpanels, overlays of theoretical curves) to make errors easier to pinpoint. As shown in Fig.~\ref{fig:correction}, the generated plot is then passed to the \textbf{Plot Judge} agent, which evaluates it against a structured rubric defined by a Pydantic schema. The rubric is domain-agnostic: fields for visual clarity, completeness, and presentation are static. The field for scientific accuracy is populated dynamically with criteria generated from the task context, including a checklist of task-specific features and their physical interpretations. The Plot Judge then returns a verdict: “continue” indicates that all criteria are satisfied and execution is handed back to Control, while “retry” signals that the plot requires revision and is accompanied by a list of observed problems. 

The \textbf{Plot Debugger} agent then ingests the VLM analysis, task context, and executed code to produce a list of targeted fixes. Its instructions emphasize tracing problems back to root causes, recognizing that multiple items on the problem list can map to a single correction. When confident, it points to specific lines and proposes explicit changes resembling a git diff; otherwise, it identifies problematic regions of the code or logic to adjust. The engineering team then receives the faulty code and the lists of problems and fixes, while all other VLM observations and the rubric itself are withheld. This begins a second pass of code generation and plot judging, which terminates either when debugging is successful or when a maximum number of loops, set as a hyperparameter, is reached.

\subsection{Steering Discovery Systems with VLM Feedback}

In discovery mode, \texttt{cmbagent} treats deviations from theoretical expectations not as errors, but as signals for further analysis. As in the correction workflow, plots are evaluated against the same structured, domain-agnostic Pydantic schema. The static fields remain unchanged, and the scientific accuracy field is again populated dynamically from the task context using GPT-4o to generate task-specific priors. These criteria capture expected features and their physical interpretations, and the \textbf{Plot Scientist} agent returns structured output with scientific observations, potential causes, signals to investigate, and a verdict of either “continue” or “explore.” This schema-based output handles routing deterministically: “continue” returns control to execution, while “explore” initiates a phase of exploratory data analysis (Fig.~\ref{fig:discovery}).

When the VLM concludes that further analysis should be added to the plan before proceeding, the \textbf{Experiment Proposer} agent generates a set of three to five candidate experiments, including the baseline, that test alternative models, parameter values, or analysis methods. Each set is accompanied by a comparison metric (e.g., $\chi^2$, WAIC, or likelihood score) that standardizes both the structure of experiments and the evaluation signal. The engineering team then runs all experiments, logging the requested metric along with additional derived values, and produces a comparison plot overlaying the results. This output is returned to the Plot Scientist, which reviews both the metrics and the visual overlays with a VLM to select a winning candidate and provide reasoning. The Experiment Proposer then synthesizes this outcome into a final task description, outlining how to update the primary analysis, what distinguishes the winning approach from the baseline, and any considerations for downstream implementation. The Plot Scientist never proposes experiments or assigns tasks to the engineer, and the Experiment Proposer never sees images, keeping the division of labor modular between visual analysis and scientific reasoning.

Throughout this workflow, predefined entries in the shared context are populated, enabling narrative generation without reconstructing the full multi-agent reasoning chain. The Experiment Proposer compresses the completed exploratory analysis into a discussion-style account that cites numerical results, explains decisions, and situates the findings within the broader research trajectory. Dual output streams ensure that discovery-related variables can be cleared without losing general execution logs, preventing stale context from propagating. Since visual inspection is separated from reasoning, the same workflow can also run in a text-only mode, where engineers output summary statistics instead of figures and the Plot Scientist acts as an LLM-as-a-judge without vision. Humans rely on plots as compact representations of complex data, but autonomous systems can generate analysis scripts at runtime, giving users direct control over whether visual evidence should guide intermediate decisions.

\section{Results}

\subsection{Case Study: Correcting CMB Power Spectra}
\label{sec:CMB}

As a demonstration of the correction workflow, we consider the task:
\begin{quote}
    Using CAMB, compute the lensed CMB temperature–temperature (TT) power spectrum with the Planck 2018 best-fit $\Lambda$CDM parameters, and plot $D_\ell = \ell(\ell+1)/(2\pi)\,C_\ell^{TT}$ for multipoles $2 \leq \ell \leq 2500$ on linear axes with units of $\mu \text{K}^2$.
\end{quote}
The initial figure produced by the engineering team was incorrect, as shown in Fig.~\ref{fig:correction}. Detecting the new plot, the Plot Judge generated expectations specific to CMB power spectra. For the first acoustic peak, it noted that theory predicts a maximum near $\ell \approx 220$ with amplitude $D_\ell \approx 5600 \, \mu K^2$. It further noted that a peak at $\ell < 200$ would imply higher matter density or faster expansion, while a peak shifted to $\ell > 220$ would suggest lower matter density or slower expansion; a depressed peak height would indicate reduced scalar amplitude or stronger damping. Similar hierarchical criteria were listed for the second and third acoustic peaks and for the high-$\ell$ damping tail.

Passing these expectations to the VLM, the Plot Judge identified several discrepancies: the peak amplitude was orders of magnitude too high, the first peak was displaced to $\ell \approx 900$, and the relative heights of subsequent peaks did not follow $\Lambda$CDM predictions. The agent returned a “retry” verdict together with a structured list of problems, triggering a handoff to the Plot Debugger. By examining the code alongside the VLM feedback, the debugger determined that the error arose from a single mistake: the CAMB function \texttt{get\_cmb\_power\_spectra()} already returns spectra scaled as $D_\ell$, but the script incorrectly applied this scaling a second time on line 29. After the engineer applied this change, the regenerated plot matched theoretical expectations. On the second pass, the Plot Judge returned a “continue” verdict, clearing the checkpoint and allowing execution to proceed.

\subsection{Case Study: Exploring Spectral Line Models}
\label{sec:CaseStudy}

As a demonstration of the discovery workflow, we consider an astrochemistry task while leaving the domain-agnostic system unchanged from the previous example. Here, the scientific challenge lies not in the theoretical difficulty of fitting spectral lines but in the time required to explore a search space of alternative models, a tedious but systematic task. In this example, we sample data from a self-absorption model but do not indicate the true distribution, saving it in a generically named file. The task is structured as follows:

\begin{quote}
\textbf{Problem Statement}: Given a new dataset, test whether the null hypothesis remains supported or should be rejected in favor of an alternative model.

\textbf{Hypothesis ($H_0$)}: The spectral line is modeled as a single Gaussian profile on a constant continuum with independent Gaussian noise: 
$$
I(v;\theta) = c_0 + A \exp\!\left[-\frac{(v - \mu)^2}{2\sigma^2}\right].
$$

\textbf{Prior Context}: Prior datasets did not provide sufficient evidence to reject $H_0$.

\textbf{New Dataset}: \texttt{path/to/data.npz} with keys “v” (velocity), “I” (intensity), and “sigma” (per-channel noise).

\textbf{Tasks}: Test $H_0$ against the new dataset. If rejected, identify and fit an alternative line-profile model.
\end{quote}

The workflow begins by fitting the null hypothesis model and plotting the result, shown in Fig.~\ref{fig:discovery}. The Plot Scientist evaluates the fit against a dynamically generated rubric. The VLM analysis highlights several discrepancies: the spectrum shows a central dip at $v \approx 0$ km/s inconsistent with a single Gaussian, the residuals form a strong “W” pattern with deviations up to $15\sigma$, and the observed dip could arise from either a self-absorption feature or a bimodal emission profile. Concluding that further exploration is required, it issues a verdict of “explore.”

Following this decision, the Experiment Proposer suggests five candidate models: the single-Gaussian baseline, two independent Gaussian emission components, a broad emission profile with a central absorbing Gaussian, a symmetric doublet, and a hybrid model with two emission components plus one absorption component. The agent specifies the Bayesian Information Criterion (BIC) as the tiebreaking evaluation metric. The engineering team runs all five experiments, returning posterior parameter estimates, fit statistics ($\chi^2$, reduced $\chi^2$, and BIC), and residual summary statistics. They also produce plots overlaying the five models on the observed data alongside a bar chart of the BIC values.

The results are passed back to the Plot Scientist, which notes that the baseline model performs extremely poorly, with a $\Delta$BIC exceeding 6000. It observes that the two-Gaussian and symmetric doublet models capture the bimodal shape but fail to reproduce the central dip, while both the self-absorption and hybrid models capture all features successfully. Reasoning that the $\Delta$BIC of 17 indicates the additional emission component in the hybrid model is not statistically justified, it selects the self-absorption model as the winner. The Experiment Proposer converts this insight into a new task for the engineering team, updates the research trajectory, and compresses the full sequence into a narrative. The engineering team then successfully fits the true distribution to the previously unknown dataset, creating the requested plot along with a subpanel of now randomly distributed residuals.

\subsection{Evaluation on Scientific Discovery Tasks}
\label{sec:evals}

\begin{table*}[t]
\centering
\caption{\textit{Benchmark results on scientific discovery tasks.} 
Reported values are pass@1 averaged across ten tasks, where success is defined as discovering the target feature in the data. 
A green \cmark\ indicates success, while a red \xmark\ indicates failure of the end-to-end workflow. 
The Coding Agent is the model responsible for code generation within the engineering team (see Fig.~\ref{fig:discovery}), and the Judge is the model powering the Plot Scientist and Experiment Proposer agents. 
``None'' indicates that no feedback loop was used. 
The spectral line case study in Sec.~\ref{sec:CaseStudy} corresponds to Q4.}

\label{tab:benchmarks}
\small
\setlength{\tabcolsep}{5pt}
\renewcommand{\arraystretch}{1.15}
\begin{tabular}{l l c c c c c c c c c c c}
\toprule
\textbf{Coding Agent} & \textbf{Judge} & Q1 & Q2 & Q3 & Q4 & Q5 & Q6 & Q7 & Q8 & Q9 & Q10 & \textbf{pass@1} \\
\midrule
Gemini 2.5 Pro                & None                & \cmark & \xmark & \xmark & \xmark & \xmark & \xmark & \xmark & \cmark & \xmark & \xmark & 0.2 \\
GPT-4.1                       & None                & \cmark & \xmark & \xmark & \xmark & \xmark & \xmark & \xmark & \xmark & \xmark & \cmark & 0.2 \\
Claude Opus 4.1               & None                & \cmark & \xmark & \xmark & \xmark & \xmark & \xmark & \xmark & \cmark & \xmark & \cmark & 0.3 \\
\texttt{cmbagent} + Gemini 2.5 Pro   & None                & \cmark & \xmark & \cmark & \cmark & \xmark & \cmark & \xmark & \cmark & \xmark & \xmark & 0.5 \\
\texttt{cmbagent} + GPT-4.1   & None                & \cmark & \xmark & \cmark & \cmark & \xmark & \xmark & \xmark & \cmark & \xmark & \xmark & 0.4 \\
\texttt{cmbagent} + GPT-4.1   & GPT-4o (LLM)        & \cmark & \xmark & \cmark & \cmark & \xmark & \cmark & \xmark & \xmark & \xmark & \cmark & 0.5 \\
\texttt{cmbagent} + GPT-4.1   & Gemini 2.5 Pro (LLM)        & \cmark & \xmark & \cmark & \cmark & \xmark & \xmark & \xmark & \cmark & \xmark & \cmark & 0.5 \\
\rowcolor{gray!20} 
\texttt{cmbagent} + GPT-4.1   & GPT-4o (VLM)        & \cmark & \cmark & \cmark & \cmark & \xmark & \cmark & \xmark & \cmark & \xmark & \cmark & 0.7 \\
\rowcolor{gray!20} 
\texttt{cmbagent} + GPT-4.1   & Gemini 2.5 Pro (VLM)& \cmark & \cmark & \cmark & \cmark & \cmark & \cmark & \xmark & \cmark & \xmark & \cmark & 0.8 \\
\bottomrule
\end{tabular}
\end{table*}

Many existing benchmarks for multi-hop reasoning in scientific problems evaluate hundreds of short tasks with discrete final answers using an LLM-as-a-judge. Exploratory work, however, involves a much larger search space and a longer task horizon, where success at the end state depends on following a rational trajectory of inductive reasoning. To evaluate systems in this setting, we construct a 10-task benchmark in which each question mirrors the case study above: a null hypothesis embedded in some scientific context, a new dataset provided without metadata or annotations, and the task of testing whether the null hypothesis should be rejected in favor of an unknown but better model. The datasets, generated by the authors, span multiple domains, including oscillators, epidemiological models, spectral line profiles, and cosmological power spectra, ensuring that the evaluation is not biased toward a single field. 

Questions 1 and 2 test the null hypothesis of a simple harmonic oscillator against data generated from a damped harmonic oscillator and a chirped harmonic oscillator, respectively. Questions 3, 4, and 5 test the null hypothesis of a single-Gaussian spectral line model against data generated from a double-Gaussian emission, emission with self-absorption, and emission with hyperfine structure. Questions 6 and 7 test the null hypothesis of a standard SEIR model against data generated from an SEIR model with waning immunity and an SEIR model with behavioral feedback. Questions 8, 9, and 10 test the null hypothesis of a $\Lambda$CDM $D_\ell$ model against data generated from a $\Lambda$CDM spectrum with an incorrect scalar spectral index, a CMB EE power spectrum, and a $\Lambda$CDM spectrum with an incorrect Hubble constant.

Performance is measured with pass@k, which gives the probability that at least one of k independent runs successfully identifies the correct underlying model. We report pass@1, which estimates the probability that a single run of the system completes the workflow correctly. Given the combinatorial space of possible model combinations and the high computational cost of each trial, this choice minimizes overall resource usage. These pass@1 results provide an initial but meaningful proof of concept; we anticipate that scaling to a larger benchmark with greater compute will enable statistically rigorous evaluation in future work.

The results are summarized in Table \ref{tab:benchmarks}. As a baseline, we evaluate the engineering team acting as the entire system. In this setup, the four coding agents already have system prompts to handle basic errors and reruns; however, there is no evaluation at inference time, as the first plot generated is treated as the final output. Performance is poor, with pass@1 between 0.2 and 0.3.

To compare the benefits of prompting agents with a workflow description versus implementing that workflow with structured outputs and specialized agents, we collapse the discovery process to rely only on \texttt{cmbagent}’s planning and control functions. In this setting, a single Researcher agent reasons over the printed statistics, assuming the roles of both the Plot Scientist and the Experiment Proposer, while the engineering team executes code. This raises pass@1 to 0.4 with GPT-4.1 as the base model, and to 0.5 when the engineers and researcher are instead instantiated with Gemini 2.5 Pro. In comparison, retaining the full structured workflow with deterministic routing, explicit schemas, and specialized agents but restricting a multimodal LLM (GPT-4o or Gemini 2.5 Pro) to text inputs for all analysis also produces a pass@1 of 0.5. In all of these text-only variants, failure modes emerge: mean posterior values may agree with theory, but subtler features such as multimodality or systematic residual patterns are often overlooked.

VLMs close this gap by evaluating plots directly and detecting problems that would otherwise require an open-ended search across a broad hypothesis space. While LLMs can flag inconsistent residual statistics, the metrics alone often fail to indicate which underlying phenomena should be probed. In contrast, a VLM can recognize the visual signature of a shifted acoustic peak or a self-absorption dip, drawing on patterns present in its training corpus. As a result, cmbagent with VLM feedback achieves pass@1 scores of 0.7–0.8 depending on the vision model, outperforming text-only variants. Beyond accuracy, VLMs also improve interpretability: visual checkpoints provide an auditable record of why a model was accepted or rejected, allowing researchers to follow the reasoning in ways that are difficult when only a sequence of statistics is available.  We theorize that the benefits of interpretability are not orthogonal to the performance improvements: figures are frequently embedded in reasoning chains that situate them within the broader research trajectory, while intermediate statistics that guide researchers toward these figures are often omitted. The statistics that are included usually serve to reinforce and ground the discussion, but they do not carry the full narrative themselves. In practice, this means text-based outputs often capture the outcome of a single analytical step, whereas plots condense the results of multiple steps into a form that directly supports scientific interpretation.

\section{Conclusion}

We introduced multi-agent systems that integrated VLMs as intermediate evaluators, extending autonomous discovery systems beyond text and code into the visual domain. By treating plots as checkpoints, we corrected errors in real time and steered exploration when unexpected features arose, enabling agents to recover from mistakes and adapt to datasets without human intervention. Case studies in cosmology and astrochemistry, supported by benchmark results, showed that VLM feedback captured features often overlooked by LLMs while producing auditable reasoning traces that statistics alone compress and obscure. These domain-agnostic workflows demonstrate that reliability and interpretability can be advanced together, moving toward agentic systems that generalize across the quantitative sciences.

\section*{Impact Statement}

We present developments to a multi-agent system for autonomous scientific research. Such systems promise scalability and utility, but responsible deployment requires deliberate safeguards: transparency and interpretability are key (see Sec.~\ref{sec:evals}). Continued progress in evaluating agentic systems is a prerequisite for their integration into research ecosystems.

\section*{Author Contributions}

KG led the integration of vision–language models, orchestration of agents with VLM feedback, and the design and execution of experiments. BB and IZ provided feedback and guidance throughout. BB led the development of \texttt{cmbagent}, including its planning and control system.

\section*{Acknowledgements}

This work was partially supported by an unrestricted gift from Google, the Cambridge Centre for Data-Driven Discovery Accelerate Programme, and the Infosys–Cambridge AI Centre. The authors are grateful to Haverford College for its support of this work and to the AG2 community.

\bibliography{paper}
\bibliographystyle{icml2025}


\newpage
\appendix
\onecolumn
\section{Benchmark Tasks}

Grouped question headings indicate tasks that share a common null hypothesis and context, while each question uses a distinct open-ended dataset to test whether the system can recognize and adapt to different features.

\begin{tcolorbox}[title={Questions 1 and 2}]
    \textbf{Problem Statement}: Given a new dataset, test whether the null hypothesis remains supported or should be rejected in favor of an alternative model.

    \textbf{Null Hypothesis ($H_0$)}: The observations are described by a simple harmonic oscillator: $$x(t;\theta) = A \cos (\omega t + \varphi) + c.$$ Phase $\varphi=0$ and offset $c=0$ are fixed, while amplitude $A$ and frequency $\omega$ are free parameters.

    \textbf{Prior Context}: Prior datasets did not provide sufficient evidence to reject $H_0$. With Dataset A (covering roughly two-thirds of a cycle, $N=25$, $\sigma=0.02$), parameter estimates were $A \approx 1.0$ and $\omega \approx 2\pi$.

    \textbf{New Dataset}: \texttt{path/to/data.npz} spans four cycles with $N = 50$, noise $\sigma = 0.08$, and keys \texttt{t} (time) and \texttt{x} (observations).
    
    \textbf{Tasks}: Test $H_0$ against the new dataset. If rejected, identify and fit an alternative model.
\end{tcolorbox}

\begin{tcolorbox}[title={Question 3}]
    \textbf{Problem Statement}: Given a new dataset, test whether the null hypothesis remains supported or should be rejected in favor of an alternative model.
    
    \textbf{Null Hypothesis ($H_0$)}: The spectral line is modeled as a single Gaussian profile on a constant continuum with independent Gaussian noise:
    $$
    I(v;\theta) = c_0 + A \exp\!\left[-\frac{(v - \mu)^2}{2\sigma^2}\right].
    $$

    \textbf{Prior Context}: Prior datasets did not provide sufficient evidence to reject $H_0$.

    \textbf{New Dataset}: \texttt{path/to/data.npz} with keys \texttt{v} (velocity), \texttt{I} (intensity), and \texttt{sigma} (per-channel noise).

    \textbf{Tasks}: Test $H_0$ against the new dataset. If rejected, identify and fit an alternative line-profile model.
\end{tcolorbox}

\begin{tcolorbox}[title={Question 4}]
    \textbf{Problem Statement}: Given a new dataset, test whether the null hypothesis remains supported or should be rejected in favor of an alternative model.
    
    \textbf{Null Hypothesis ($H_0$)}: The spectral line is modeled as a single Gaussian profile on a constant continuum with independent Gaussian noise:
    $$
    I(v;\theta) = c_0 + A \exp\!\left[-\frac{(v - \mu)^2}{2\sigma^2}\right].
    $$

    The centroid parameter $\mu$ is fixed at $0$, corresponding to the systemic velocity.
    
    \textbf{Prior Context}: Prior datasets did not provide sufficient evidence to reject $H_0$.

    \textbf{New Dataset}: \texttt{path/to/data.npz} with keys \texttt{v} (velocity), \texttt{I} (intensity), and \texttt{sigma} (per-channel noise).

    \textbf{Tasks}: Test $H_0$ against the new dataset. If rejected, identify and fit an alternative line-profile model.
\end{tcolorbox}

\begin{tcolorbox}[title={Question 5}]
    \textbf{Problem Statement}: Given a new dataset, test whether the null hypothesis remains supported or should be rejected in favor of an alternative model.
    
    \textbf{Null Hypothesis ($H_0$)}: The HCN spectral line is modeled as a single Gaussian profile on a constant continuum with independent Gaussian noise:
    $$
    I(v;\theta) = c_0 + A \exp\!\left[-\frac{(v - \mu)^2}{2\sigma^2}\right].
    $$

    \textbf{Prior Context}: Prior datasets did not provide sufficient evidence to reject $H_0$.

    \textbf{New Dataset}: \texttt{path/to/data.npz} with keys \texttt{v} (velocity), \texttt{I} (intensity), and \texttt{sigma} (per-channel noise).
    
    \textbf{Tasks}: Test $H_0$ against the new dataset. If rejected, identify and fit an alternative line-profile model.
\end{tcolorbox}

\begin{tcolorbox}[title={Questions 6 and 7}]
    \textbf{Problem Statement}: Given a new dataset, test whether the null hypothesis remains supported or should be rejected in favor of an alternative model.
    
    \textbf{Null Hypothesis ($H_0$)}: The outbreak dynamics follow a standard SEIR model for active infections $I(t)$ under homogeneous mixing and constant parameters:
    \begin{align*}
    \frac{dS}{dt} &= -\beta S I, \\
    \frac{dE}{dt} &=  \beta S I - \sigma E, \\
    \frac{dI}{dt} &=  \sigma E - \gamma I, \\
    \frac{dR}{dt} &=  \gamma I.
    \end{align*}

    \textbf{Prior Context}: Prior datasets did not provide sufficient evidence to reject $H_0$.

    \textbf{New Dataset}: \texttt{path/to/data.npz} with keys \texttt{t} (time), \texttt{I} (active infections), and \texttt{sigma\_I} (measurement uncertainty).

    \textbf{Tasks}: Test $H_0$ against the new dataset. If rejected, identify and fit an alternative epidemiological model.
\end{tcolorbox}

\begin{tcolorbox}[title={Questions 8, 9, and 10}]
    \textbf{Problem Statement}: Given a new dataset, test whether the null hypothesis remains supported or should be rejected in favor of an alternative model.

    \textbf{Null Hypothesis ($H_0$)}: The observed values are drawn from the same statistical distribution as the lensed CMB TT power spectrum predicted by $\Lambda$CDM with Planck 2018 parameters.

    \textbf{Prior Context}: Previous analyses used \texttt{CAMB} to compute the CMB power spectra and \texttt{emcee} for MCMC sampling.

    \textbf{New Dataset}: \texttt{path/to/data.npz} with keys \texttt{ell} (multipole $\ell$), \texttt{Dl\_obs} (TT bandpowers in $\mu\mathrm{K}^2$), and \texttt{sigma\_Dl} (uncertainties in $\mu\mathrm{K}^2$).

    \textbf{Tasks}: Test $H_0$ against the new dataset and plot the results. If rejected, propose and fit an alternative model that better explains the data.
\end{tcolorbox}

\section{Prompt Templates}

\begin{tcolorbox}[title={Plot Judge Prompt}]
    You are a plot judge analyzing a scientific plot. Your task is to evaluate the plot's quality and provide structured feedback.
    Analyze this plot across scientific accuracy, visual clarity, completeness, and professional presentation.
    
    \medskip
    Context about the goal of the plot: \{improved\_main\_task\}

    \medskip
    IMPORTANT: We do not use LaTeX rendering at all in plots, so do not ask for it or comment on unrendered TeX code.

    \medskip
    Request plot elements that are scientifically beneficial for understanding the plot. 
    Don't request them just for rubric completeness if the plot is clear without them. 
    Don't request additional annotations.

    \medskip
    Be thorough and critical - the plot will only be accepted if ALL criteria are met. 
    If any criterion is not satisfied, list the specific problems found in the problems field.

    \medskip
    Your verdict must be either "continue" (plot fully meets all criteria) or "retry" (plot needs improvements).
\end{tcolorbox}

\begin{tcolorbox}[title={Plot Debugger Prompt}]
    You are a plot debugging expert. The VLM has identified problems with a plot, and you need to provide targeted code fixes.

    \medskip
    TASK CONTEXT: \{task\_context\}

    \medskip
    VLM ANALYSIS: \{vlm\_analysis\}

    \medskip
    PROBLEMS IDENTIFIED: \{'\textbackslash n'.join(f"- {p}" for p in problems)\}

    \medskip
    EXECUTED CODE: \{executed\_code\}

    \medskip
    YOUR JOB: Analyze the code and provide targeted fixes for each problem. Multiple problems can often be caused by a single underlying issue in the code, so focus on root causes rather than symptoms.

    \medskip
    For each fix:
    
    - If you can identify specific code lines/sections: Reference them directly (e.g., \texttt{Line 15: change plt.xlim() to...})

    - If the issue is broader: Provide helpful considerations for the engineer
    
    - Be concrete and actionable, focusing on code changes needed to address the visual/scientific issues
    
    - Group related problems into single fixes when appropriate
\end{tcolorbox}

\begin{tcolorbox}[title={Experiment Finalization Prompt}]
You are creating the final implementation task for an engineer based on experiment comparison results. Convert the winning approach into actionable engineering instructions.

\medskip
ORIGINAL TASK: \{task\_context\}

\medskip
ORIGINAL CODE: \{original\_code\}

\medskip
VLM WINNER SELECTION: \{winner\_selection\}

\medskip
VLM REASONING: \{winner\_reasoning\}

\medskip
FULL VLM ANALYSIS: \{vlm\_analysis\}

\medskip
WINNING EXPERIMENT DETAILS: \{winning\_experiment\}

\medskip
ALL EXPERIMENTS TESTED: \{proposed\_experiments\}

\medskip
Your job is to create a complete final task description that:

1. Provides a clear, actionable description of what the engineer should implement using the winning approach

2. Includes concrete guidance on parameters, methods, and code modifications that are needed

3. Defines success criteria that indicate whether the implementation is correct

4. Explains how the winning approach differs from the original baseline

\medskip
If the winning approach is the original baseline, the final task should instruct the engineer to continue with the original method but apply any improvements suggested during analysis.

\medskip
The final description must be actionable and specific enough for an engineer to implement immediately.
\end{tcolorbox}

\begin{tcolorbox}[title={Plot Scientist Experiment Evaluation Prompt}]
    You are a senior scientist evaluating experiment comparison results. Analyze the comparison plot showing multiple experimental approaches and select the winning method.

    \medskip
    ORIGINAL TASK: \{task\_context\}

    \medskip
    EXPERIMENTS TESTED: \{experiments\_info\}

    \medskip
    RESULTS SUMMARY: \{results\_info\}

    \medskip
    COMPARISON METRIC: \{comparison\_metric\}

    \medskip
    YOUR TASK: 
    
    1. Analyze the comparison plot showing all experimental approaches
    
    2. Compare performance based on the \{comparison\_metric\} values
    
    3. Select the experiment that performed best overall (including "Original" as an option)
    
    4. Provide clear scientific reasoning for your selection
    
    5. Explain the performance differences between approaches

    \medskip
    EVALUATION CRITERIA:
    
    - Raw \{comparison\_metric\} performance (consider whether higher or lower values are better)
    
    - Magnitude of improvement over baseline
    
    - Scientific validity and robustness
    
    - Practical interpretability

    \medskip
    IMPORTANT: The "Original" baseline approach is always a valid choice if the new experiments did not provide clear improvements.

    \medskip
    Analyze the comparison plot carefully and make your selection based on scientific merit.
\end{tcolorbox}

\begin{tcolorbox}[title={Scientific Criteria Prompt (Correction Mode)}]
You are a scientific expert analyzing plots. Generate domain-specific scientific accuracy criteria for error detection.

\medskip
Context: \{plot\_description\}

\medskip
Your response will be used as criteria in a VLM prompt, so be direct and specific. Do not include conversational phrases.

\medskip
First, identify key features that should have specific expected coordinates/values (x-axis, y-axis positions, ratios, etc.). For each feature, specify:

1. Expected x/y coordinates or values

2. What deviations indicate and why they are scientifically invalid

3. What physical processes cause these features

\medskip
IMPORTANT: Only include features you are confident about. Skip any where the expected values can vary significantly or you are uncertain. It is better to have fewer, more reliable criteria than many uncertain ones.

\medskip
Example format:

"Feature name: Expected at x $\approx$ [value], y $\approx$ [value]

- If shifted to x $<$ [value]: indicates [physical cause] (invalid because [reason])

- If shifted to x $>$ [value]: indicates [physical cause] (invalid because [reason])

- If shifted to y $<$ [value]: indicates [physical cause] (invalid because [reason])"

\medskip
Example (stellar main sequence):

"Main sequence turnoff: Expected at B-V $\approx$ 0.6, $M_V$ $\approx$ 4.0 for solar metallicity

- If shifted bluer (B-V $<$ 0.4): indicates higher metallicity/younger age (invalid for old globular clusters)

- If shifted redder (B-V $>$ 0.8): indicates lower metallicity/older age (invalid for young open clusters)"

\medskip
Provide similar specific criteria for this plot type, focusing only on features with well-defined expected values.

\medskip
There are cases where specific values are not known beforehand or not the focus of the plot. When this is the case, provide other distinct and discrete features that are required to be present.

\medskip
Example (exoplanet transit light curve):

Check the following features:

- Phase of transit dip: should be at phase = 0.0

\hspace{0.3cm}- If not at phase = 0.0 → indicates wrong timing or ephemeris (invalid for phased data)
    
- Shape of transit dip: should be symmetric and flat-bottomed

\hspace{0.3cm}- If not symmetric/flat-bottomed → indicates wrong planet/star size or reduction error (invalid for known system)
    
- Depth of transit dip: should reach normalized flux $\approx$ 0.99 (1 percent depth)

\hspace{0.3cm}- If depth is incorrect → indicates wrong planet/star size or reduction error (invalid for known system)
    
- Ingress and egress: should have similar duration

\hspace{0.3cm}- If durations differ → may indicate reduction error or systematics
    
- Baseline flux: should be flat at flux = 1.0 before and after transit

\hspace{0.3cm}- If baseline is not flat → indicates improper normalization or stellar variability (invalid for detrended light curves)
\end{tcolorbox}

\begin{tcolorbox}[title={Experiment Proposer Prompt}]
You are designing a comprehensive scientific experiment comparing multiple approaches. Based on VLM observations, propose 3--5 structured experiments that can be implemented and compared systematically.

\medskip
TASK CONTEXT: \{task\_context\}

\medskip
VLM SCIENTIFIC OBSERVATIONS: \{observations\_str\}

\medskip
POTENTIAL CAUSES TO INVESTIGATE: \{causes\_str\}

\medskip
SIGNALS OR REGIONS TO EXAMINE: \{signals\_str\}

\medskip
CURRENT BASELINE CODE: \{executed\_code\}

\medskip
FULL VLM ANALYSIS: \{vlm\_analysis\}

\medskip
Design experiments that:

1. Include the original baseline approach as experiment number 1 for comparison

2. Test alternative models, free parameters, or analysis methods

3. Are clearly differentiated and scientifically motivated

4. Can be implemented by an engineer and compared using a single metric

5. Apply wide ranges in priors when unknown

6. Probe for new discoveries, not statistical relaxations that optimize for metrics

\medskip
Choose one quantitative metric that can be calculated for all experiments.

\medskip
COMPARISON VISUALIZATION STRATEGY:

- Default approach: One plot showing all experiments with their metric values on a bar chart, table, or comparison plot

- When comparing models to data: Use subplots with a metric comparison on the left and all model overlays on the right

- When comparing theoretical curves: Use subplots with a metric comparison on the left and theory versus data overlays on the right

- Goal: One final comparison visualization that clearly shows all metric values

\medskip
Each experiment must include clear implementation guidance for the engineer.

\medskip
Implementation hints must instruct the engineer to print quantitative metrics in a clear, organized format. Requirements:

- Group results by experiment with clear headers such as "Experiment 1: Name"

- Include quantitative results such as parameter values, fit statistics, and performance metrics

- End with a summary comparing all experiments using the chosen metric

- Use consistent formatting that makes it easy to identify which results belong to which experiment
\end{tcolorbox}

\begin{tcolorbox}[title={Scientific Criteria Prompt (Discovery Mode)}]
    You are a scientific expert analyzing plots. Generate domain-specific discovery criteria for identifying scientifically interesting patterns that warrant further investigation.

    \medskip
    Context: \{plot\_description\}

    \medskip
    Your response will be used as criteria in a VLM prompt, so be direct and specific. Do not include conversational phrases.

    \medskip
    Based on this specific scientific context, identify domain-specific patterns that would be particularly noteworthy for this type of analysis. Focus on what would be unexpected, anomalous, or scientifically significant for this particular field, measurement, or experiment. Provide specific criteria that go beyond the general discovery framework already established.
\end{tcolorbox}

\begin{tcolorbox}[title={Exploratory Experiment Prompt}]
You are a scientific experiment designer. Based on VLM observations of a plot, propose specific and actionable experiments or analyses that can be implemented in code.

\medskip
TASK CONTEXT: \{task\_context\}

\medskip
VLM SCIENTIFIC OBSERVATIONS: \{observations\_str\}

\medskip
POTENTIAL CAUSES TO INVESTIGATE: \{causes\_str\}

\medskip
SIGNALS OR REGIONS TO EXAMINE: \{signals\_str\}

\medskip
CURRENT CODE: \{executed\_code\}

\medskip
FULL VLM ANALYSIS: \{vlm\_analysis\}

\medskip
Based on these scientific observations, propose specific experimental suggestions that:

1. Are actionable and implementable in code

2. Help investigate observed patterns or anomalies

3. Provide statistical validation or refutation of the observations

4. Could reveal underlying causes of the patterns

5. Are grounded in the scientific context of the task

\medskip
Focus on concrete experiments such as parameter sweeps, statistical tests, model comparisons, sensitivity analyses, or data subsampling. The suggestions must be specific enough for an engineer to implement directly.
\end{tcolorbox}

\begin{tcolorbox}[title={Scientific Narrative Prompt}]
You are a senior scientist writing a scientific discovery narrative. Tell the story of what was discovered, using specific numerical results.

\medskip
ORIGINAL SCIENTIFIC TASK: \{task\_context\}

\medskip
VISION-LANGUAGE MODEL ANALYSIS OF ORIGINAL PLOT

\medskip
Scientific Observations Identified: \{observations\_text\}

\medskip
Potential Causes Hypothesized: \{causes\_text\}

\medskip
Signals or Regions Flagged for Investigation: \{signals\_text\}

\medskip
Proposed Experiments for Investigation: \{experiments\_summary\}

\medskip
Comparison Metric Selected: \{comparison\_metric\}

\medskip
EXPERIMENT EXECUTION OUTPUT AND METRICS: \{experiment\_execution\_output\}

\medskip
EVALUATION AND WINNER SELECTION

\medskip
VLM Experiment Analysis: \{vlm\_experiment\_analysis\}

\medskip
VLM Metric Comparison: \{vlm\_metric\_comparison\}

\medskip
Winner Selected: \{vlm\_winner\_selection\}

\medskip
VLM Winner Reasoning: \{vlm\_winner\_reasoning\}

\medskip
VLM Performance Summary: \{vlm\_performance\_summary\}

\medskip
FINAL IMPLEMENTATION

\medskip
New Primary Task Description: \{final\_task\_description\}

\medskip
Key Differences from Original Approach: \{final\_differences\_from\_original\}

\medskip
YOUR TASK:

Write a structured scientific discovery narrative with exactly five labeled sections. Each section must be three to six sentences long and must include specific numerical results from the experiment execution output.

\medskip
REQUIRED STRUCTURE:

1. INITIAL SETUP: Describe the original goal, the baseline approach, and expectations.

2. DISCOVERY MOMENT: Summarize the anomalies flagged by the VLM and why they merited further investigation.

3. INVESTIGATION: Describe the experiments that were run and the comparison metric used.

4. REALIZATION: Explain what the winning experiment revealed, including specific numerical values from the experiment execution output.

5. UPDATED UNDERSTANDING: State the new scientific conclusion and what it reveals about the underlying physical system.

\medskip
The narrative must reference multiple numerical values from the experiment execution output and must be written as a scientist explaining to colleagues what was discovered.
\end{tcolorbox}


\end{document}